\documentclass[letterpaper]{article} 
\usepackage{aaai24}  
\usepackage{times}  
\usepackage{helvet}  
\usepackage{courier}  
\usepackage[hyphens]{url}  
\usepackage{graphicx} 
\usepackage{natbib}  
\usepackage{caption} 
\usepackage{algorithm}
\usepackage{algorithmic}
\usepackage{amsfonts}
\usepackage{multirow}
\usepackage{subcaption}
\usepackage{xcolor}
\usepackage{pifont}
\usepackage{amsmath}
\usepackage{newfloat}
\usepackage{listings}
\DeclareCaptionStyle{ruled}{labelfont=normalfont,labelsep=colon,strut=off} 
\floatstyle{ruled}
\newfloat{listing}{tb}{lst}{}
\floatname{listing}{Listing}
\title{Multi-Scene Generalized Trajectory Global Graph Solver with Composite Nodes for Multiple Object Tracking}
\author{
    Yan Gao\textsuperscript{\rm 1},
    Haojun Xu\textsuperscript{\rm 1},
    Nannan Wang\textsuperscript{\rm 1},
    Jie Li\textsuperscript{\rm 1$*$},
    Xinbo Gao\textsuperscript{\rm 1,\rm 2}\thanks{Corresponding Author.}
}
\affiliations{
    \textsuperscript{\rm 1}Xidian University\\
    \textsuperscript{\rm 2}Chongqing University of Posts and Telecommunications\\

    \{yangao, haojunxu\}@stu.xidian.edu.cn, 
    nnwang@xidian.edu.cn, 
    leejie@mail.xidian.edu.cn, 
    gaoxb@cqupt.edu.cn
}

\usepackage{bibentry}

\begin{document}

\maketitle

\begin{abstract}
	The global multi-object tracking (MOT) system can consider interaction, occlusion, and other ``visual blur'' scenarios to ensure effective object tracking in long videos. Among them, graph-based tracking-by-detection paradigms achieve surprising performance. However, their fully-connected nature poses storage space requirements that challenge algorithm handling long videos. Currently, commonly used methods are still generated trajectories by building one-forward associations across frames. Such matches produced under the guidance of first-order similarity information may not be optimal from a longer-time perspective. Moreover, they often lack an end-to-end scheme for correcting mismatches. This paper proposes the Composite Node Message Passing Network (CoNo-Link), a multi-scene generalized framework for modeling ultra-long frames information for association. CoNo-Link's solution is a low-storage overhead method for building constrained connected graphs. In addition to the previous method of treating objects as nodes, the network innovatively treats object trajectories as nodes for information interaction, improving the graph neural network's feature representation capability. Specifically, we formulate the graph-building problem as a top-k selection task for some reliable objects or trajectories. Our model can learn better predictions on longer-time scales by adding composite nodes. As a result, our method outperforms the state-of-the-art in several commonly used datasets.
\end{abstract}

\section{Introduction}
Multi-Object Tracking (MOT) aims to generate trajectories for all moving objects in a video stream. It is a fundamental module for video content analysis in application areas such as autonomous driving and intelligent robotics. In recent years, tracking-by-detection as the dominant paradigm in the field divides the task into (i) frame-by-frame object detection and (ii) data association, i.e., linking potential targets to the correct object trajectory whenever possible.
\begin{figure}[h]
	\centering
	\includegraphics[width=0.85\linewidth]{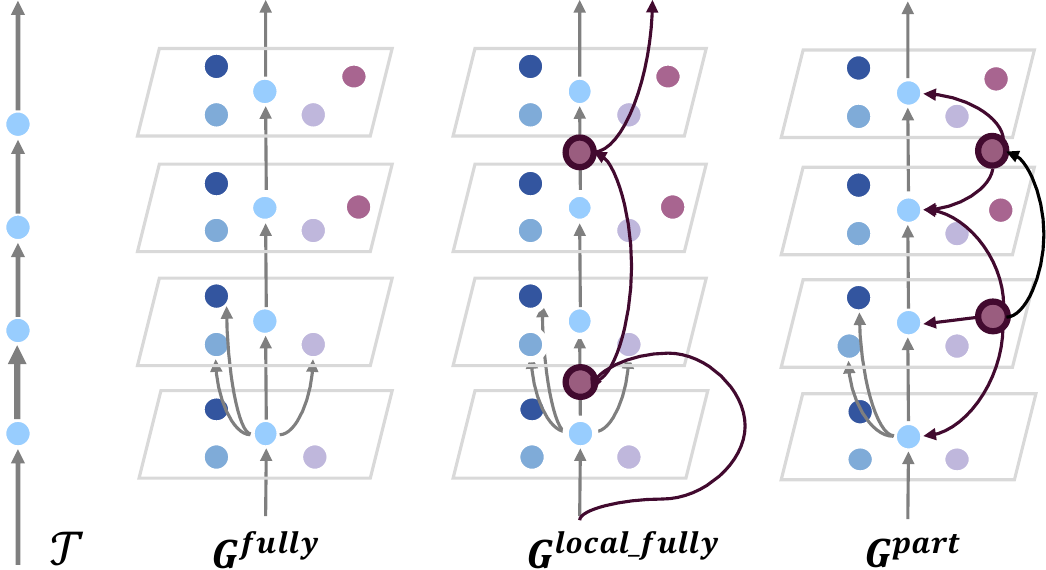}
	\caption{Node connectivity relationship graphs for different graph structures. The 2nd bolded gray link in $ T $ indicates that the connection in each graph is the relationship between the blue node in the first frame and the nodes in the next frame.}
	\label{fig:concept_map}
\end{figure}
Data association is performed mainly on neighboring frames, greedily matching trajectories with detection proposals through simple cues such as position and motion prediction \cite{eccv_ZhangSJYWYLLW22_ByteTrack,iccv_Tokmakov0BG21_PermaTrack,eccv_ZhouKK20_CenterTrack,iccv_BergmannML19_Tracktor,icip_BewleyGORU16_SORT} or appearance features \cite{cvpr_PangQLCLDY21_QDTrack,eccv_WangZLLW20_JDE,ijcv_ZhangWWZL21_FairMOT}. The trajectories formed by such local trackers can satisfy the accuracy requirements. However, severe object occlusion and appearance changes in crowded scenes can pose challenges for long-term identity preservation.
Many studies have executed offline associations over the entire video frame to obtain long-range trajectories. Such global trackers typically require the construction of global appearance and motion cues or pairwise association between tracklets on all frames based on graphical representations.

To maximize the applicability to each scenario, recent graph-based MOT studies \cite{cvpr_BrasoL20_MPNTrack,icml_HornakovaHRS20_Lif_T,iccv_HornakovaKSRRH21_ApLift,cvpr2023_Cetintas_SUSHI} have adopted a generic approach to global tracking by using a unified module to process videos in a single framework. While the research has shown promising results, there are still some issues with the generalized graph-based approach. A significant challenge is the need to consider the time cues of video clips more. Generic methods generally use a uniform, fully-connected graph-building system, and due to storage constraints, the node information encoded is limited to neighboring frames, forming trajectories that may not be optimal in the temporal dimension. In contrast, when humans perform dynamic tracking, they must maintain the temporal consistency of the perceived object \cite{Hyvrinen2014CurrentUO}. Another issue is that the fixed mode of building graphs suffers from an error accumulation effect. The incorrectly predicted features can ultimately affect the quality of feature learning.
The above analysis raises the question, what does the graph structure needed for a generic MOT approach look like? 

In this work, we propose CoNo-Link, a composite node message-passing network that combines improved graph formulation rules and effective learning strategies into a graph solver for unified learning. The framework comprises a NodeNet that generates nodes to build a graph and a graph neural network (GNN) link block that accomplishes reasoning about the entire domain. Specifically, NodeNet inputs long clips and outputs a one-piece partially connected graph $ G^{part} $ containing potential connections and composite nodes. We then reason by the link block to predict the edge score for $ G^{part} $ and output the final result.

$ G^{part} $ is the core of our CoNo-Link, inspired by the human visual system. Research \cite{Hyvrinen2014CurrentUO} shows that infants can expand from tracking salient objects to arbitrary objects when they observe the world. That means observation and tracking is a continuous process in which there are priorities of interest. We summarize three tracking characteristics in the visual system and design two types of nodes and three types of connections accordingly. Among them, the detection node priority confirms the significantly moving objects, the trajectory node records the continuous motion of things, and the connections between composite nodes create pathways for the expression of relationships between arbitrary objects. We use a GNN to process the trajectory nodes and propose a complementary edge learning strategy for these nodes that allows potentially misclassified edges to resume learning.  

Fig. \ref*{fig:concept_map} includes the node connectivity relations of different graph structures when generating a trajectory $ T $. In the view of blue nodes, both the fully-connected frame-by-frame graph $ G^{fully} $ and the hierarchical neighboring frames graph $ G^{local\_fully} $ have to be connected to all the nodes of the next frame, while our $ G^{part} $ only needs to be connected to meaningful object nodes. This scheme solves two limitations of the previous approaches: (i) Partially link significantly reduces memory consumption and can efficiently handle long clips. (ii) Distant nodes are reachable in $ G^{part} $, with better feature learning potential. We evaluate the proposed CoNo-Link on several benchmarks: achieve a competitive 83.7 IDF1 and 82.7 MOTA on MOT17, 81.8 and 77.5 on MOT20, and 64.1 and 89.7 on DanceTrack.

In summary, we make the following main contributions.
\begin{itemize}
	\item We propose a composite node messaging network that achieves globally optimal convergence by gradually aggregating meaningful trajectory nodes. It guarantees continuous solutions with significantly lower computational costs.
	\item We construct a partially connected graph based on composite nodes to build a perfect graph domain in which a generalized MOT graph solver can effectively utilize temporal information. It is a natural graph structure abstracted from human visual tracking.
	\item We achieve the state-of-the-art result in three public datasets and have multi-scene applicability.
\end{itemize}

\section{Related Work}
\label{related work}
\subsection{Local Short-Term Tracking.}
Many modern trackers run online using frame-by-frame or local frames \cite{icip_BewleyGORU16_SORT,iccv_BergmannML19_Tracktor,cvpr_MeinhardtKLF22_TrackFormer,cvpr_PangQLCLDY21_QDTrack,iccv_Tokmakov0BG21_PermaTrack,eccv2022_Zeng_MOTR,eccv_ZhangSJYWYLLW22_ByteTrack,ijcv_ZhangWWZL21_FairMOT,tip_GaoXZLG22_OPITrack}. Distance metrics for motion and spatial cues are often central to the design of trajectory formation for these methods. SORT \cite{icip_BewleyGORU16_SORT} and DeepSORT \cite{icip_WojkeBP17_DeepSORT,wacv_WojkeB18_CosineML} have led many methods to use the Kalman filter as motion models \cite{icip_BewleyGORU16_SORT,eccv_ZhangSJYWYLLW22_ByteTrack,eccv_WangZLLW20_JDE,ijcv_ZhangWWZL21_FairMOT,nips_DendorferYOL22_QuoVadis}. CenterTrack \cite{eccv_ZhouKK20_CenterTrack} implements motion prediction and execution association through neural networks, which has led to the emergence of methods to improve motion prediction through improved model structures. Tracktor \cite{iccv_BergmannML19_Tracktor} inspired a frame-by-frame regression-based tracking framework \cite{AAAI_LiangZZLH22,cvpr_MeinhardtKLF22_TrackFormer,eccv_ZhouKK20_CenterTrack,iccv_BergmannML19_Tracktor}. Some trackers use appearance to identify the same object to increase robustness in low-quality video scenes \cite{cvpr_PangQLCLDY21_QDTrack,ijcv_ZhangWWZL21_FairMOT,iccv_Xu0ZH19_RelationTrack,icip_WojkeBP17_DeepSORT}, such as low frame rates or strong camera motion \cite{nips_DendorferYOL22_QuoVadis}. Although these pair-wise association-based trackers have good tracking stability, they do not focus on the long-term preservation of object identity. Here we do this by performing association on all objects throughout the temporal dimension.

\subsection{Graph-based global tracking.}
Graphs are a framework well suited for modeling data association, where each object trajectory can be considered a simple graph with entry and exit degrees of 1 for each node except for the start and end nodes. They use object detection as object nodes and represent edges as possible trajectory hypotheses. In contrast to trackers that use neighbor frame information, graph-based approaches define the cross-frame object association problem as a global combinatorial optimization problem \cite{AAAI_KohKYKKC22,Laura_MOT20,cvpr_BrasoL20_MPNTrack,eccv2022_Zeng_MOTR}. To this end, many studies have used different optimization strategies, including multi-cuts \cite{cvpr_TangAAS17_Multicut}, minimal cliques \cite{eccv_ZamirDS12_GMCP}, network flow \cite{pami_BerclazFTF11_KShort,cvpr_ButtC13_MinCost}, and disjoint path approaches \cite{icml_HornakovaHRS20_Lif_T,iccv_HornakovaKSRRH21_ApLift}. Following \cite{cvpr_BrasoL20_MPNTrack}, this paper relies on a simplified minimum cost flow formulation \cite{cvpr_ZhangLN08_GANF} to ensure that graph-based network structures are manageable.

Early graph-based methods often used methods based on handcrafted models \cite{cvpr_TakalaP07_handdesign} or conditional random fields \cite{cvpr_YangN12a_CRF} to obtain association cues. More recently, many approaches employ GNNs \cite{ijcv_BrasoCL22_NMP,cvpr_BrasoL20_MPNTrack,cvpr_DaiWCZHD21_LPC} or Transformers \cite{cvpr_MeinhardtKLF22_TrackFormer,cvpr_ZhouYKK22_GTR,eccv2022_Zeng_MOTR} to learn features for the association. Lif\_T \cite{icml_HornakovaHRS20_Lif_T} incorporates epistemic and pose features in optimizing disjoint path problems. MPNTrack \cite{cvpr_BrasoL20_MPNTrack} proposes a neural solver to optimize a simplified graph. MOTR \cite{eccv2022_Zeng_MOTR} follows the DETR \cite{eccv_CarionMSUKZ20_DETR} structure to iteratively update the tracking queries in a propagated manner. The global association method GTR \cite{cvpr_ZhouYKK22_GTR} uses transformer queries to generate the entire trajectory simultaneously. These GNN and transformer-based works have achieved good performance. However, the current class of methods iteratively completes object building through all frames and still suffers from the limitations of local information association. In addition, out-of-memory (OOM) is prone to occur if the length of the processed video is considerable.

Recently multi-stage tracking methods \cite{cvpr_Chen0HW20_MGL,cvpr_Wu2021_TraDes} become popular. These methods first form short-range trajectories and then deals with lost tracklets and occlusions over long periods. Different optimization techniques \cite{cvpr_Wu2021_TraDes,icra_KW21_JDEGNN} and cues \cite{cvpr_GuptaDG19_LVIS} are designed for multiple-stage merging when forming long-range trajectories. Many modern trackers incorporate such techniques to improve error correction \cite{pami_BerclazFTF11_KShort,cvpr_BrasoL20_MPNTrack,cvpr_MeinhardtKLF22_TrackFormer,wacv_WojkeB18_CosineML}.
In this paper, we propose a learning strategy to gain the ability to reload missing edges during training.

\section{Methodology}
\label{headings}
\subsection{Preliminaries}
\label{subsec:preliminaries}
\subsubsection{Tracking-by-Detection (TBD).} 
The TBD paradigm performs the task through frame-by-frame object detection and well-designed inter-target association. 
Let $ I $ be a set of images in a video; the detector first identifies and locates targets in all elements of $ I $, producing a set of objects $ O $ with positions \{$ p_i $ $\vert$ $ p_i \in \mathbb{R}^4 $\}. The object set $ O $ used for the association is determined in some way, and it is common practice to keep the proposals above a set confidence threshold as candidate objects. Each candidate object $ o_i $ contains the position $ p_i $, the image patch corresponding to $ p_i $, and the time $ t $. Then, the tracker associates each object in $ O $ to obtain its trajectory over time $ T = \{ \tau_{1}, \tau_{2}, \cdots, \tau_{K} \} $, and $ \tau_{i} = \{ {o_i}_1, {o_i}_2, \cdots, {o_i}_{n_i} \} $ where $ n_i $ is the trajectory length of the object $ o_i $. In this paper, our algorithm also follows this classical paradigm.

Our tracker is built on modeling inter-frame object relations using an undirected graph $ G = (V, E) $ where each node $ V $ of $ G $ corresponds to an object detection $ o_i $. The edge $ E \to \mathbb{R}^{V \times V} $ denotes the possible interactions between objects on different frames. The association assumption $ E $ guarantees the connectivity of object pairs in different frames and facilitates trajectory error correction.

\subsubsection{Graph-Based Tracking Baseline.}
Based on $ G $ as the underlying representation of inter-object relations, the proposed model is performed on a message-passing graph network tracker \cite{cvpr_BrasoL20_MPNTrack} based on the classical MOT network flow formulation \cite{cvpr_ZhangLN08_GloAssNetFlow}. It is a framework that performs global reasoning directly over the entire graph domain and predicts whether object connectivity relations hold. Formally, it represents a trajectory $ \tau_i $ as an object $ o_i $ in $ G $ linked time-by-time, i.e., $ E(\tau_i) = {(o_{i1}, o_{i2}), ..., (o_{in_{i-1}},o_{in_i})} $. Then, the corresponding paths in $ G $ are subjected to binary decision through discrete optimization. Thus, if edge $ (u,v) \to E(\tau_i) $ denotes a correct assumption, it means that the edge cost is $ y(u,v) = 1 $; otherwise, it is an incorrect assumption, i.e., $ y(u,v) = 0 $. The algorithm obtains the final trajectory by classifying the edges in the graph.
\begin{figure}[h]
	\centering
	\includegraphics[width=\linewidth]{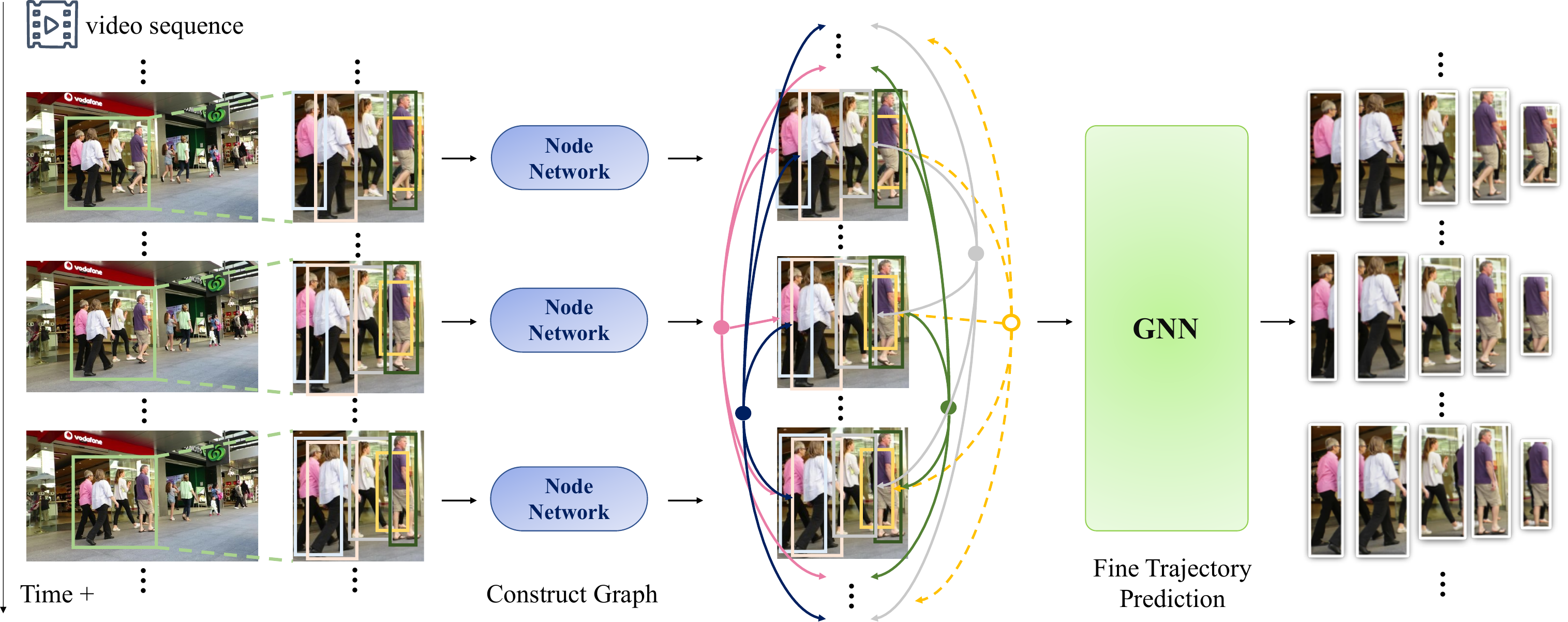}
	\caption{We propose the CoNo-Link framework. It comprises a node network (NodeNet) and a graph neural network (GNN). Given long-time video clips as inputs, the NodeNet first performs a graph-building operation to construct the objects appearing throughout the clips into graph nodes. Then, the constrained connection graph of the clip completes the prediction of delicate trajectories in the massage passing GNN.}
	\label{fig:pipline}
\end{figure}

\subsection{CoNo-Link}
\subsubsection{Tracking in A Gradual Way.} We provide an overview of CoNo-Link in Fig. \ref{fig:pipline}. It comprises a node network (NodeNet) and message-passing graph neural networks (GNN). The NodeNet first performs the preparatory work of constructing the constrained connectivity graph in a video clip. Specifically, it constructs the objects appearing throughout the clip as graph nodes. In this, we define the graph nodes as composite node representations that contain the object's detection nodes $ N^{det} $ and trajectory nodes $ N^{traj} $ (with an initial length of 1). Then, we establish the connections between composite nodes $ \{N^{det}, N^{traj}\} $ according to the graph-building rules in the next section. Eventually, the complete partial connection graph of the entire input video clip will be fed into the GNN to predict delicate trajectories. It is worth noting that the NodeNet is trained separately. The trajectory nodes built by it are rough representations containing global information, which is intended to ensure the envelopment of edges included in the built graph to the ground truth edges. In addition, CoNo-Link can uniformly access all the information in long video clips, thus ensuring that object identities can be aligned over a wide range of time scales.

Our CoNo-Link is a tracking algorithm that gradually learns the connectivity between composite nodes through GNN, i.e., an incremental prediction process from detection nodes to sub-trajectory nodes and then from sub-trajectory nodes to the complete trajectory. This progressive learning reduces the learning difficulty of the GNN, which can effectively deal with ultra-long-time partially connected graphs and thus accomplish long-term tracking.

\begin{figure}[h]
	\centering
	\includegraphics[width=\linewidth]{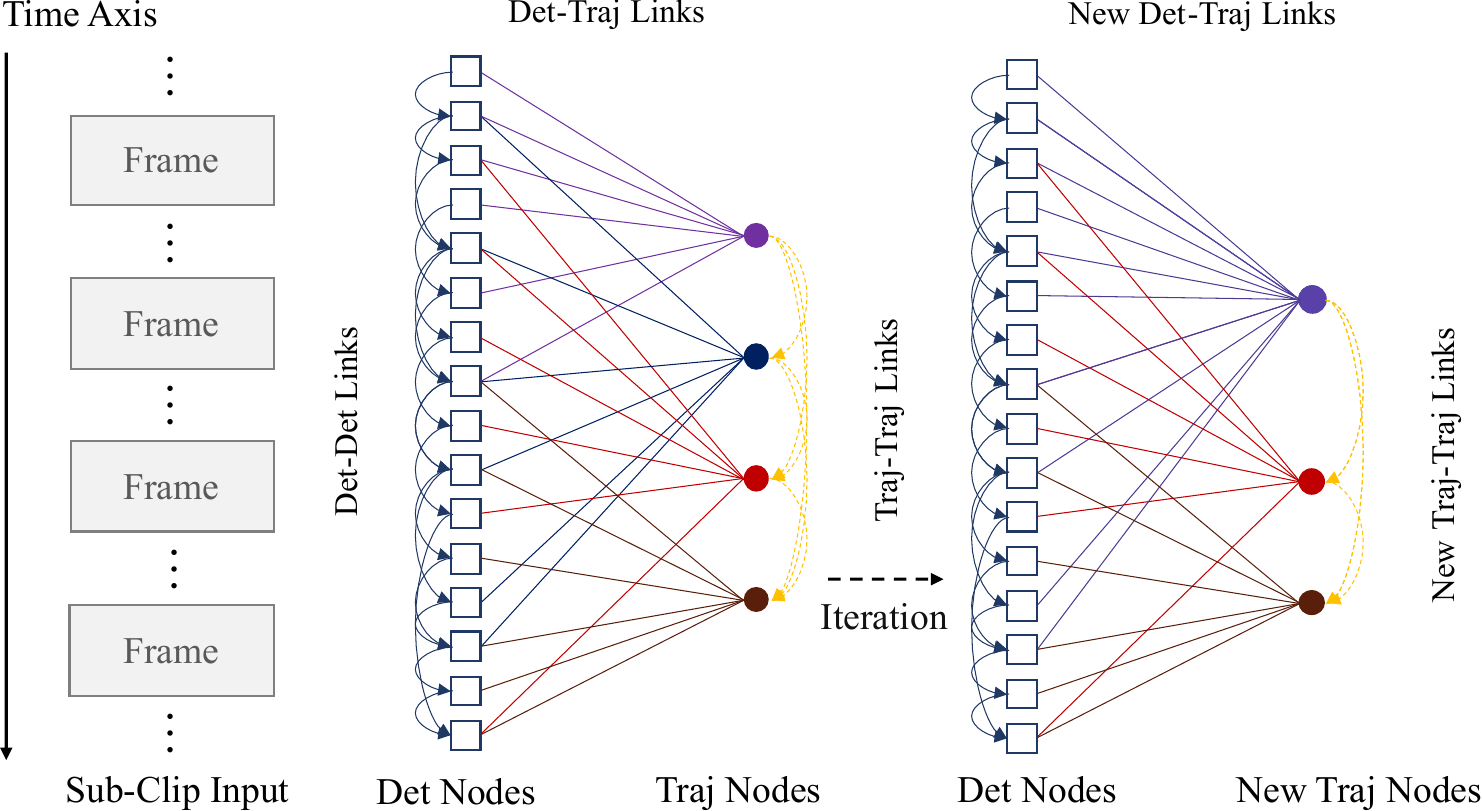}
	\caption{$ G^{part} $ structure. This graph contains potential links with a high ground-truth link share, which is the GNN input.}
	\label{fig:part_graph}
\end{figure}

\subsubsection{Constructing A Partially Connected Tracking Graph.}
\label{sec:graph_build}
This study proposes matching using a composite node-based partial connectivity graph $ G^{part} $ learning global information interactions over the time series of the entire video clip. Based on our experiments, we found that generating a fully connected graph and pruning it will always miss the edges. There are two types of reasons; one is that the same identity (ID) target is not in the video sampling window, and the other is that the connection is missing due to the feature learning problem of the model itself. Since missing links are unavoidable, we abandon the commonly used pruning method and instead construct the graph by identifying the ``most promising'' node relationships and backfilling the edges. Among them, the nodes can be the correct object detection or the shortest paths with the same ID that is easy to construct. The main idea is to follow a coarse-to-fine strategy: build an imperfect but relatively correct part of the node relationship graph and then determine new connections from the GNN to realize accurate tracking. Therefore, we propose NodeNet to establish the initial relationships between detections.

\begin{figure*}[h]
	\centering
	\includegraphics[width=0.9\linewidth]{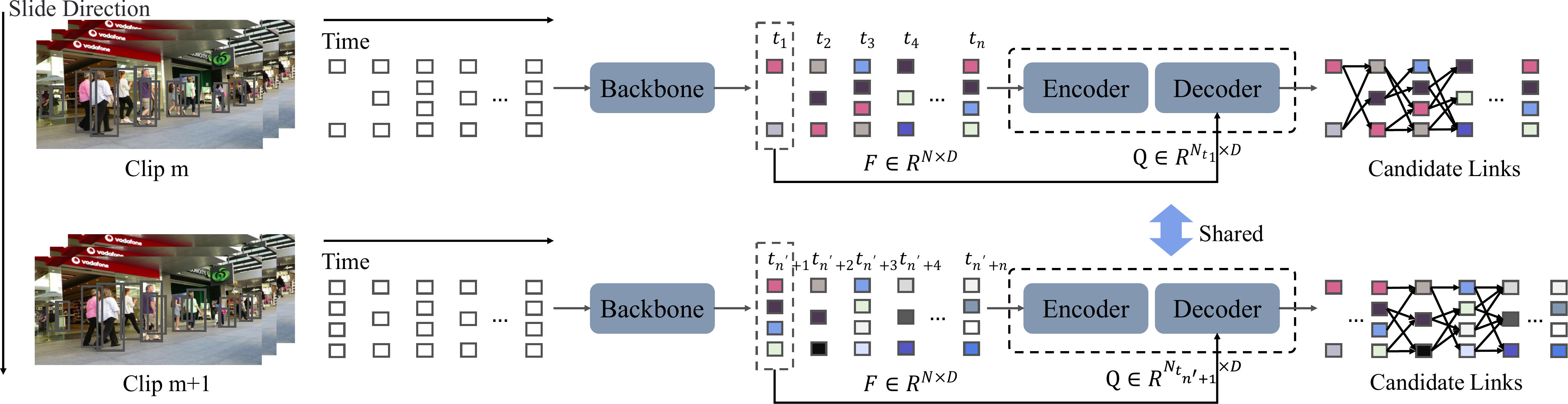}
	\caption{The detailed network architecture of NodeNet. It contains a layer of an encoder and a layer of a decoder following \cite{eccv_CarionMSUKZ20_DETR,cvpr_ZhouYKK22_GTR}. The object feature F and the query Q are the inputs to the encoder and decoder, respectively, and generate an affinity matrix between the query and the object $ \mathcal{M} $ as the basis for developing our candidate links.}
	\label{fig:node_net}
\end{figure*}

\textit{Establish The Initial Relationships.} We employ a query representation to capture objects' temporal and interaction relations. Specifically, we give each query node an initially constructed coarse trajectory in the semantic feature space by  track search but keep only the candidate links for each node along the temporal dimension. Let $ O = \{o^t_1, \cdots, o^t_{N_t}\} $ be a set of detector output objects for image $ I^t $ with total number $ N = \sum^T_t N_t $. Let $ F^t = \{f^t_i \to R^D | f^t_1, \cdots, f^t_{N_t}\} $ be the set of $ D $-dimensional features extracted from their corresponding bounding boxes and $ F = F^1 \cup \cdots \cup F^T $ is the set of all features in time slice $ T $. As shown in Fig. \ref{fig:node_net}, all objects with $ F \in R^{N \times D} $ and query $ Q_k \in R^D $ are the inputs to our NodeNet. The association scores $ s \in R^N $ that produce the coarse trajectories are the outputs. Formally, we use a softmax activation to model the likelihood of association between an object $ i $ and each trajectory $ k $ at time $ t $ as $ P_M(o_i|Q_k,F)=exp(s^t_i)/\sum_{j \in \{\emptyset, 1, \cdots, N_t\}} exp(s^t_j) $. Then, the distribution of all objects at time $ t $ corresponding to trajectory $ k $ is $ P^t (p|Q_k,F) = \sum^{N_t}_{i=1} l_{[p=p^t_i]} P_M(o_i|Q_k,F) $, where $ l_{[\cdot]} $ assigns a bounding box $ p $ to each query. Ultimately, the distribution of trajectory $ k $ over the entire period is $ P^T (\tau|Q_k,F) = \prod^T_{t=1} P^t(\tau^t|Q_k,F) $. 

During training, we maximize the log-likelihood of the ground-truth trajectory, and features that are not matched are used as background queries and supervised empty set frames. Let $ \hat o_k $ be matched objects for a ground-truth trajectory $ \tau $, and $ F_{\hat o_k}^m $ be the features of $ \hat o_k $. For each trajectory $ \tau $, we optimize the training objective of the query assignment as:
\begin{footnotesize}
	\begin{equation}
		\begin{split}
			& L_\text{NodeNet} = L_\text{bg}(F) + \sum_{\tau} L_\text{matched}(F, \tau). \\
			& L_\text{bg}(F)   = - \sum\limits_{m = 1}^T \sum\limits_{j: {\exists\mkern-9mu/} \hat o_k^m = j}\sum\limits_{t = 1}^T \log {P_M}(o^t=\emptyset|F_j^m,F). \\
			& L_\text{matched}(F,\tau) = - \sum\limits_{\tiny{m \in \{ 1.... .T|\hat o_k^m \ne \phi \}}}{\sum\limits_{t = 1}^T {\log {P_M}(\hat o_k^m|F_{\hat o_k^m}^m,F)}}.
		\end{split}
	\end{equation}
\end{footnotesize}

We first obtain the similarity matrix $ \mathcal{M} $ in the inference process. The method is to let NodeNet process the video stream as a sliding window. The length of the video stream is $ T_{clip} = 512 $, the window size is $ T = 32 $, and the step size is 16.
We use a backbone network to extract the detected frames within the window to obtain $ F $ sequentially. Then, let  $ Q = K = V = F $ feed them into NodeNet, and output an association matrix of $ N \times N $. Assign this matrix to the corresponding position in the large matrix $ \mathcal{M} $, and if there is already a value in the corresponding place, take the average value in the overlapping part.

After obtaining the similarity matrix $ \mathcal{M} $, we will perform a frame-by-frame association. For the first frame, we initialize all detections to trajectories. For the $ N_t $ bounding boxes in the subsequent frame $ I^t $, the similarity sub-matrix $ M $ of the bounding boxes within the window of size $ min(t, T) $ is taken out in $ \mathcal{M} $ with this frame as the end of the window. The similarity between the query detections (trajectories) and the object detections within the window that have the same ID is summed to obtain the similarity matrix of each ID to the object of the current frame $ \bar{M} \in R^{{ID} \times N_t}$. In addition, we compute the IoU matrix $ \hat{M} \in R^{{ID} \times N_t } $ between the last occurrence of each ID box and the current frame box. Our final cost matrix is $ C = - max(\bar{M}, \hat{M}) $. Further, we use the Hungarian algorithm \cite{books_Kuhn10_Hungarian} to get the optimal association for each ID. We start a new trajectory if the average association score with any previous ID is below a threshold $\vartheta$. Otherwise, we attach the generated trajectory's underlying current detection (query) to the matching existing trajectory. In the similarity matrix $ \bar{M} $, we connect the optimal associations for each ID and the Top-$ k $ bounding boxes with the highest similarity. These detections connections (\textit{Det-Det links}) are the potential links we need.

\textit{Predict Several Short Paths.} We feed $ G^{part} $ containing $ T_{clip} $ information into the GNN for edge prediction. Then, we aggregate some natural short trajectories generated by the GNN into trajectory nodes $ N^{traj} $. Precisely, we fill in the predicted scores of the GNN into $ \mathcal{M}_{GNN} $ as the similarity between the nodes and determine the trajectory IDs using the same association mode for generating Det-Det links. For each path (containing Det-Traj links), the detection features within the trajectory are averaged as the new trajectory feature representation. Finally, we build connections between $ N^{traj} $ (Traj-Traj links) and do not establish the relationship if temporal IoU exists between $ N^{traj} $.

$ G^{part} $ is now the entire form that the complete set of candidate links established between pre-generated composite nodes. It contains Det-Det links, Det-Traj links, and Traj-Traj links, and we illustrate its construction in Fig. \ref{fig:part_graph}.
These three links are significant because the Det-Det link generates new $ N^{traj} $. Det-Traj link is responsible for building the information interaction between $ N^{det} $ and $ N^{traj} $. The Traj-Traj link is responsible for fusing the sub-trajectories. In each iteration, the composite nodes undergo a directed transformation, and we transform the Traj-Traj links that satisfy the requirements into new $ N^{traj} $, which are then merged with the existing $ N^{traj} $ to generate the final result.

\textit{Summary of $ G^{part} $ vs. $ G^{fully} $.} The pre-generated composite nodes retain meaningful object nodes and reduce invalid edges, alleviating the labeling imbalance. This strategy reduces storage occupancy, so the graph contains all the information in clip time, guaranteeing spatial information interaction over long distances. At the same time, establishing multiple rather than a single kind of edges makes some nodes unavailable to reachable, making mega-graph long-distance matching plausible. Through upper-bound analysis experiments, we determined that the restricted connected graph based on composite node learning performs better than the hierarchical neighboring frame fully connected graph learning. Also, proof that our method satisfies globally optimal matching is provided in the supplementary.

\subsubsection{Learn An Effective Link Tracker.}
We use a message-passing GNN \cite{cvpr_BrasoL20_MPNTrack} to process a partially connected graph with composite nodes. Our main contribution is a learning strategy for GNN training. Specifically, the Traj-Traj links of our $ G^{part} $ are fully-connected edges constructed for $ N^{traj} $. Trajectory nodes are relatively few at this time, from 10k down to around the 0.1k level (see Table \ref{tab:upper_bound}), making it possible to build fully-connected graphs between them, and can minimize the problem of missing links when creating graphs. We obtain the final trajectory by performing edge classification to decide whether to keep the established connections. We show the learning process in algorithm \ref{alg:algorithm}. The key to making the GNN learn efficiently is maintaining the proportion of positive sample edges, which is why our method works. In the following, we detail a GNN link block that makes long-time node connections.
\begin{algorithm}[tb]
	\caption{Graph aggregation algorithm}
	\label{alg:algorithm}
	\textbf{Input}: current graph node $V$, current graph edge $E^I$\\
	\textbf{Parameter}: GNN $f_{\theta_1}, f_{\theta_2}$, and $ \theta_1 = \theta_2 $, assign threshold $\epsilon$\\
	\textbf{Output}: id of every node $K$
	\begin{algorithmic}[1] 
		\STATE Compute edge classification score $S_{E^I} = f_{\theta_1}(V, E^I)$
		\STATE Assign id to every node in $V$ by $S_{E^I}$ use the tracker described in Sec 3.2.1. The id of $V$ is $K_0$.
		\STATE Aggregate the same id node in $V$ to get trajectory node $V_\text{node}$
		\STATE Get fully-connected edge $E_\text{fc}$ of $V_\text{node}$
		\STATE Compute edge classification score $S_{E_\text{fc}} = f_{\theta_2}(V_\text{node}, E_\text{fc})$
		\STATE $E_+ = E_\text{fc}[S_{E_\text{fc}} > \epsilon]$
		\STATE Find the connected and nontime-nonoverlap node group
		\STATE Give the same id to every node in node group and map the id in $V_\text{node}$ to $V$. The id of $V$ now is $K$.
		\STATE \textbf{return} $K$
	\end{algorithmic}
\end{algorithm}

\textit{Link Block.}
Let $ G^I = (V^I, E^I) $ be a graph of our iteration $ I $. $ h^{(0)}_n $ is the node embedding for each $ n \in V^I $ and $ h^{(0)}_{(u,v)} $ is the edge embedding for each $ (u,v) \in E^I $. Following the time-aware framework of \cite{cvpr_BrasoL20_MPNTrack}, the GNN aims to learn a function that encodes the higher-order semantic context contained in the node and edge feature vectors via information propagation. It feeds the edge embedding to an multi-layer perceptron (MLP) and outputs a score $ y^{(u,v)}_{pred} = MLP(h^{(s)}_{(u,v)}) $, where $ (s) $ denotes the number of message passes. We set this score to a similarity with a minimum threshold $ \epsilon $ limitation for association. As mentioned, we compute similarity scores based on association cues between nodes recorded in $ \mathcal{M}_{GNN} $. The cues contain the similarity of the appearance embedding and the estimated score based on the closest temporal distance. Additional  details are provided in the supplementary material.

\subsubsection{Training.} CoNo-Link contains two message-passing GNNs \cite{cvpr_BrasoL20_MPNTrack} which are trained jointly at all levels. For this purpose, we first train the first network and add the second network to train at a certain level. This ensures stability during the training process. Specifically, we unfreeze the second network after 5000 training iterations. We trained them using a focal loss \cite{focal_loss_pami20} with a $ \gamma = 1 $ and summed all the losses as our final loss.

\section{Experiments}
\subsection{Datasets and Metrics.}
We conducted experiments on three public benchmarks: MOT17 \cite{MOT15_arxiv15,MOT16_arxiv16}, MOT20 \cite{MOT20_arxiv20}, and DanceTrack \cite{dancetrack_cvpr22}. The MOT series is a dense pedestrian tracking dataset evaluated under public and private detection protocols. DanceTrack is a dataset of dance videos with similar appearance and complex motion patterns. We evaluate the performance of the algorithms using several widely used metrics: IDF1 \cite{IDF1_eccvw16}, MOTA \cite{MOTA_tpami09}, and HOTA \cite{HOTA_ijcv21}. IDF1 focuses on identity maintenance quality, while the official metric MOTA focuses on detection quality and tracking stability (IDS). HOTA combines two aspects to unify detection localization and association performance.  

\subsection{Implementation details}
Training parameters: we used the pre-trained ResNet50-IBN \cite{cvpr_DaiWCZHD21_LPC} as our ReID network and froze it during training. The GNNs were co-trained with a learning rate of $ 3\times10^{-4} $, weight decay of $ 10^{-4} $, and a batch of 2 clips with 200 epochs. The optimizer was Adam \cite{ADAM_iclr15}. Inference: our model can handle sequences of arbitrary length, and for training and memory efficiency, we have $ T_{clip} $ of 512 frames. During inference, we fill the trajectory gaps by linear interpolation. The runtime on MOT17 is 19 FPS with given detections. Object detection: for DanceTrack and the private settings of MOT17 and MOT20, we follow the detections obtained in the YOLOX \cite{YOLOX_arxiv21} trained in \cite{eccv_ZhangSJYWYLLW22_ByteTrack}. Our tracker was implemented on a single RTX 3090 GPU using Python 3.10 and Pytorch 2.0.0 \cite{pytorch_nips19}.

\subsection{Ablation Study}
\subsubsection{Experimental settings.} In this section, we use the MOT17 dataset for all experiments. To evaluate our model, we employ four video sequences (04, 05, 09, and 11) for training and the most challenging three sequences (02, 10, and 13) for validation. All ablation experiments were performed on the validation set.

\subsubsection{Candidate Link Generation Structure.} 
\begin{figure}[h]
	\centering
	\includegraphics[width=1\linewidth]{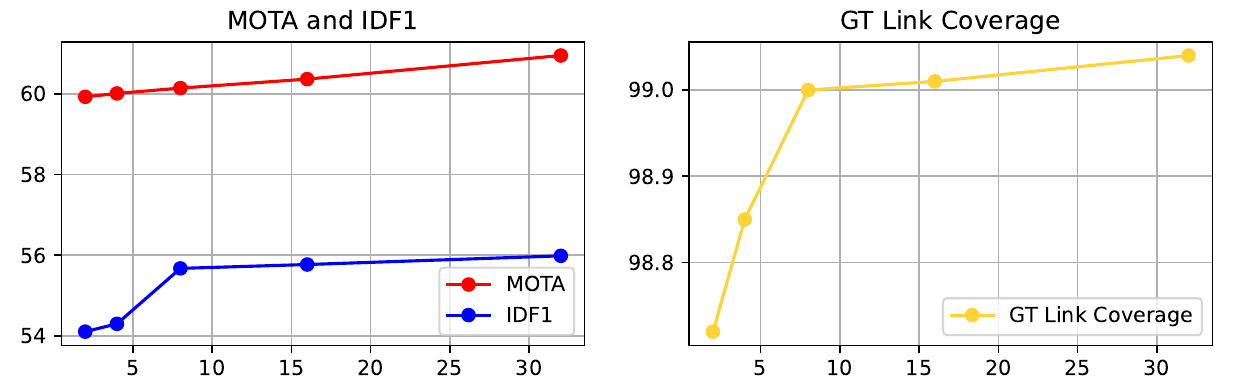}
	\caption{The window size experiment of NodeNet.}
	\label{fig:nodenet_T}
\end{figure}

We consider NodeNet's three patterns, IoU, ReID, and Trans, for generating potential connections. IoU and ReID denote Hungarian matching \cite{books_Kuhn10_Hungarian} based on IoU and ReID cues, and Trans. denotes query-based Transformer matching (i.e., for similarity associations considering global information). Based on these three models, we start the search for the relation-building range Top-$ k $ using MOTA, IDF1, and the ground truth (GT) links coverage  (Cover.) as evaluation metrics. To obtain as high edge coverage as possible without being affected by environmental factors, we use GT boxes as inputs to NodeNet based on the purity assumption. Table \ref{tab:nodenet_structure} shows the results. We use query matching to establish relationships between the top 5 matches. Although $ k = 10 $ performs best, the gain is small, and the number of edges becomes large. We also show in Fig. \ref{fig:nodenet_T} the effect of the window size for establishing potential connections on these metrics. There is the highest GT envelope at $ T = 32 $ and does not has memory overflow.
\begin{table}[!t]
	\centering
	\scalebox{0.7}{
		\begin{tabular}{llllllllll}
			\hline
			\multicolumn{4}{c}{MOTA $ \uparrow $}  & \multicolumn{3}{c}{IDF1 $\uparrow$} & \multicolumn{3}{c}{Cover. $\uparrow$} \\
			\hline
			Top-$ k $ & IoU & ReID & Trans. & IoU & ReID & Trans. & IoU & ReID & Trans. \\
			\hline
			1 & 59.3 & 18.3 & 60.8 & 48.8 & 2.9 & 55.4 & 95.8 & 49.9 & 98.7 \\
			2 & 59.9 & 28.7 & 60.9 & 50.1 & 4.4 & 55.9 & 97.3 & 75.2 & 99.0 \\
			5 & 60.0 & 43.7 & \bf 61.0 & 50.3 & 9.8 & \bf 56.0 & 97.6 & 95.0 & 99.0 \\
			10 & 60.0 & 52.5 & \bf 61.0 & 50.3 & 18.8 & \bf 56.0 & 97.8 & 98.8 & \bf 99.1 \\
			\hline
	\end{tabular}}
	\caption{Candidate link generation method selection.}
	\label{tab:nodenet_structure}
\end{table}
\subsubsection{Temporally Connected Graph Domains.} We investigate the upper bounds on the performance of our proposed partially connected graph and the recent hierarchical fully connected graph. The experiment still uses GT detection as input for graph construction. We observe higher upper bounds for almost all metrics of $ G^{part} $ in Table \ref{tab:upper_bound}, including a reduction of IDS to 18. We calculate the edge and node numbers in Table \ref{tab:cca} for both graphs under the same upper bound. $ G^{local\_fully} $ has an average of 256,095 edges and 7,482 nodes, whereas our $ G^{part} $ has only 143,603 edges and 151 nodes. The results show that the proposed $ G^{part} $ achieves 99.9\% GT edge coverage before rounding than $ G^{local\_fully} $ when using fewer edges to build the graph. As expected, our $ G^{part} $ is compelling in graph domain construction because it preserves temporal connectivity in addition to considering neighboring information. It allows the GNN to ``see'' nodes over time and make more credible decisions, thus generating significantly less IDS.
\begin{table}[!t]
	\centering
	\scalebox{0.7}{
		\begin{tabular}{llllllll}
			\hline
			\# & Graph & MOTA $ \uparrow $  & IDF1 $\uparrow$ & IDS $\downarrow$ & HOTA $\uparrow$ & MT $\uparrow$ & Cover. $\uparrow$ \\
			\hline
			1 & $ G^{local\_fully} $ & 61.7 & 76.2 & 38 & 61.4 & 315 & \bf 99.9 \\
			2 & $ G^{part} $ & \bf 61.8 & \bf 76.3 & \bf 18 & \bf 61.5 & \bf 316 & \bf 99.9 \\
			\hline
	\end{tabular}}
	\caption{Experiments on the upper bound of graph structures.}
	\label{tab:upper_bound}
\end{table}
\subsubsection{CoNo-Link's Component Ablation Experiments.} We show the performance of each part on the MOT17 validation set in Table \ref{tab:abl}. The first row shows the results after feeding $ G^{part} $ directly into the GNN. Then, we add the proposed learning strategy (LS) to get the results of the Link blocks in rows 2-5. There are four settings for LS: (i) mapping the valid edges ($ \epsilon > 0.5 $) in Traj-Traj links to the corresponding Det-Det links and tracking the Det-Det links to get the final results; (ii) mapping the Traj-Traj links and their classification scores to the corresponding Det-Det links, and track the Det-Det links to get the final results. (iii) Map all the Traj-Traj links to the corresponding Det-Det links and use GNN to re-extract the scores of the corresponding edges and track to output the final results. (iv) Use Traj-Traj links to directly fuse $ N^{traj} $ to obtain the final results. Except for LS\_1, which slightly decreases MOTA, the other strategies improve in all metrics. In particular, our LS\_4 has 0.2 \% of MOTA, 11.8 \% of IDF1, and 187 IDS gains. Note that all results in Table \ref{tab:abl} are without interpolation.
\begin{table}[!t]
	\centering
	\scalebox{0.75}{
		\begin{tabular}{lllll}
			\hline
			\# & Method & MOTA $ \uparrow $  & IDF1 $\uparrow$ & IDS $\downarrow$ \\
			\hline
			1& $ G^{fully} $ + GNN & 59.6 & 57.8 & 844 \\
			2& $ G^{part} $ + GNN & 60.1 (0.5 $ \uparrow $) & 58.1 (0.3 $ \uparrow $) & 636 (208 $ \downarrow $) \\
			3& $ G^{part} $ + GNN + LS\_1 & 59.9 (0.2 $ \downarrow $) & 64.4 (6.3 $ \uparrow $) & 553 (83 $ \downarrow $) \\
			4& $ G^{part} $ + GNN + LS\_2 & 60.1 (0.0 -) & 61.3 (3.2 $ \uparrow $) & 601 (35 $ \downarrow $) \\
			5& $ G^{part} $ + GNN + LS\_3 & 60.2 (0.1 $ \uparrow $) & 65.2 (7.1 $ \uparrow $) & 534 (102 $ \downarrow $) \\
			6& $ G^{part} $ + GNN + LS\_4 & \bf 60.3 (0.2 $ \uparrow $) & \bf 69.9 (11.8 $ \uparrow $) & \bf 449 (187 $ \downarrow $)\\
			\hline
	\end{tabular}}
	\caption{Ablation study of CoNo-Link.}
	\label{tab:abl}
\end{table}

\begin{table}[!t]
	\centering
	\scalebox{0.7}{
		\begin{tabular}{llccccc}
			\hline
			\# & Method     & FLOPs (G)  & FPS & VRAM Footprint (MiB) & $ E_{avg} $ & $ N_{avg} $ \\
			\hline 
			1 & $ G^{local\_fully} $  & 46.7 & 17.1 & 3,689 & 256,095 & 7,482 \\
			2 & $ G^{part} $  & \bf 36.3 & \bf 18.9 & \bf 3,523 & \bf 143,603 & \bf 151 \\
			\hline
	\end{tabular}}
	\caption{Computational complexity experiments.}
	\label{tab:cca}
\end{table}

\subsection{Benchmark Results}
\subsubsection{MOT.} 
We report the results of our model in Table~\ref{tab:mot17+20} for MOT17 and MOT20 under the private detection protocol. Our approach achieves the desired results on both challenges. On MOT17, we outperform SUSHI based on hierarchical fully-connected graphs on all metrics. In the highly crowded scene of MOT20, it slightly underperforms SUSHI in identity switches. This result may be due to the large number of relational interactions in a short period that impact our approach. Still, again, the global information ensures that there are enough accurate tracklets in the results. Compared to ByteTrack, our model improves this by 6.4 IDF1, 4.0 HOTA, and 2.4 MOTA, reducing IDS by 50.3\%. Our performance demonstrates the positive significance of the time-domain connectivity partially connected graph.

\subsubsection{DanceTrack.} Table~\ref{tab:dance} demonstrates that in scenarios with multiple complex motion patterns, we have good improvements on all metrics and that our approach has a better balance between IDF1 and MOTA performance without excessive loss of accuracy. Results of this dataset show that an overall evaluation of trajectories in the temporal dimension can provide some correct clues for association.

\begin{table}[!t]
	\centering
	\scalebox{0.65}{
		\begin{tabular}{lllll}
			\hline
			Method     & IDF1 $\uparrow$ & HOTA $\uparrow$ & MOTA $ \uparrow $ & IDS $\downarrow$ \\
			\hline
			\multicolumn{5}{c}{MOT17}\\
			\hline
			MPNTrack$\star$ \cite{cvpr_BrasoL20_MPNTrack} & 61.7 & 49.0 & 58.8 & 1185 \\
			QDTrack \cite{cvpr_PangQLCLDY21_QDTrack}           & 66.3 & 53.9 & 68.7 & 3378 \\
			TrackFormer \cite{cvpr_MeinhardtKLF22_TrackFormer} & 68.0 & 57.3 & 74.1 & 2829 \\
			MOTR \cite{eccv2022_Zeng_MOTR}                     & 68.6 & 57.8 & 73.4 & 2439 \\ 
			PermaTrack \cite{iccv_Tokmakov0BG21_PermaTrack}    & 68.9 & 55.5 & 73.8 & 3699 \\ 
			MeMOT \cite{cvpr_CaiX0XXTS22_MeMOT}                & 69.0 & 56.9 & 72.5 & 2724 \\ 
			GTR \cite{cvpr_ZhouYKK22_GTR}                      & 71.5 & 59.1 & 75.3 & 2859 \\ 
			FairMOT \cite{ijcv_ZhangWWZL21_FairMOT}            & 72.3 & 59.3 & 73.7 & 3303 \\ 
			GRTU \cite{iccv_Wang21_GRTU}                       & 75.0 & 62.0 & 74.9 & 1812 \\ 
			CorrTracker \cite{cvpr_WangZPX21_CorrTracker}      & 73.6 & 60.7 & 76.5 & 3369 \\ 
			Unicorn \cite{eccv_YanJSWYLL22_Unicorn}            & 75.5 & 61.7 & 77.2 & 5379 \\ 
			ByteTrack \cite{eccv_ZhangSJYWYLLW22_ByteTrack}    & 77.3 & 63.1 & 80.3 & 2196 \\
			SUSHI \cite{cvpr2023_Cetintas_SUSHI} & 83.1 & 66.5 & 81.1 & 1149 \\
			\hline
			Ours       & \bf 83.7 & \bf 67.1 & \bf 82.7 & \bf 1092 \\ 
			\hline
			\multicolumn{5}{c}{MOT20}\\
			\hline
			MPNTrack$\star$ \cite{cvpr_BrasoL20_MPNTrack} & 59.1 & 46.8 & 57.6 & 1210 \\
			TrackFormer \cite{cvpr_MeinhardtKLF22_TrackFormer} & 65.7 & 54.7 & 68.6 & 1532 \\ 
			MeMOT \cite{cvpr_CaiX0XXTS22_MeMOT}                & 66.1 & 54.1 & 63.7 & 1938 \\ 
			FairMOT \citealp{ijcv_ZhangWWZL21_FairMOT}         & 67.3 & 54.6 & 61.8 & 5243 \\ 
			GSDT \cite{icra_KW21_JDEGNN}                       & 67.5 & 53.6 & 67.1 & 3131 \\ 
			CorrTracker \citealp{cvpr_WangZPX21_CorrTracker}   & 69.1 & –    & 65.2 & 5183 \\ 
			ByteTrack \cite{eccv_ZhangSJYWYLLW22_ByteTrack}    & 75.2 & 61.3 & 77.8 & 1223 \\
			SUSHI \cite{cvpr2023_Cetintas_SUSHI}      & 79.8 & 64.3 & 74.3 & \bf 706  \\ 
			\hline
			Ours             & \bf 81.8 & \bf 65.9 & \bf 77.5 & 956 \\ 
			\hline
	\end{tabular}}
	\caption{Test set results on MOT17 and MOT20 benchmark. MPNTrack$\star$ is the result under the public protocol.}
	\label{tab:mot17+20}
\end{table}

\begin{table}[t]
	\centering
	\scalebox{0.6}{
		\begin{tabular}{llllll}
			\hline
			Method       & IDF1 $\uparrow$ & HOTA $\uparrow$ & MOTA $ \uparrow $ & AssA $\uparrow$ & DetA $\uparrow$ \\
			\hline
			CenterTrack [69] & 35.7 & 41.8 & 86.8 & 22.6 & 78.1 \\ 
			FairMOT \cite{ijcv_ZhangWWZL21_FairMOT}     & 40.8 & 39.7 & 82.2 & 23.8 & 66.7 \\ 
			TraDes \cite{cvpr_Wu2021_TraDes}      & 41.2 & 43.3 & 86.2 & 25.4 & 74.5 \\ 
			GTR \cite{cvpr_ZhouYKK22_GTR}         & 50.3 & 48.0 & 84.7 & 31.9 & 72.5 \\ 
			QDTrack \cite{cvpr_PangQLCLDY21_QDTrack}     & 50.4 & 54.2 & 87.7 & 36.8 & 80.1 \\ 
			MOTR \cite{eccv2022_Zeng_MOTR}        & 51.5 & 54.2 & 79.7 & 40.2 & 73.5 \\ 
			ByteTrack \cite{eccv_ZhangSJYWYLLW22_ByteTrack}   & 53.9 & 47.7 & 89.6 & 32.1 & 71.0 \\ 
			SUSHI \cite{cvpr2023_Cetintas_SUSHI}     & 63.4 & 63.3 & 88.7 & 50.1 & 80.1 \\
			\hline
			Ours             & \bf 64.1 & \bf 63.8 & \bf 89.7 & \bf 50.7 & \bf 80.2 \\
			\hline
	\end{tabular}}
	\caption{Test set results on DanceTrack benchmark.}
	\label{tab:dance}
\end{table}

\section{Conclusion}
This paper presents CoNo-Link, a feasible method for oversized graph tracking. Through ablation experiments, we demonstrate that partially connected graphs can reduce the video memory footprint and efficiently handle long video clips; the trajectory learning strategy can improve the upper bound of $ G^{part} $ performance. In addition, the proposed method outperforms the state-of-the-art approaches on three benchmarks. In the future, we would like to design an interactive approach with CLIP \cite{CLIP_ICML21} to enhance the graph solver's potential in tracking through multimodal tasks, e.g., image caption \cite{image_captioning_aaai20}.

\section{Acknowledgments}
This work is supported in part by the National Natural Science Foundation of China under Grants 62176195, 62036007, U22A2096, U21A20514, and 62221005; in part by the Technology Innovation Leading Program of Shaanxi under Grant 2022QFY01-15; in part by Open Research Projects of Zhejiang Lab under Grant 2021KG0AB01.

\bibliography{aaai24}

\clearpage

{--Supplementary Material--}

\begin{abstract}
	In this material, we provide:
	(i) a more detailed explanation of our approach's central ideas (Section \ref{sec:proof}A).
	(ii) Comparison with other methods on MOT datasets under public protocols (Section \ref{sec:benchmark_results}B).
	(iii) a description of the details of our implementation (Section \ref{sec:other_details}C). 
	(iv) details of the structure (Section \ref{sec:architecture}D).
	and (v) a demonstration of the visualization resulting from our approach (Section \ref{sec:visualization}E).
\end{abstract}

\section{A. Central Ideas}
\label{sec:proof}
\subsubsection{Borrow Talent from Predecessor.} This paper is built on the shoulders of giants \cite{cvpr_BrasoL20_MPNTrack} in exploring graph solvers. To obtain a subgraph partition of the trajectory of the graph $ \mathcal{G} = (\mathcal{V}, \mathcal{E}) $, we approximate the set of ground truth edges by the edge prediction $ \bar{y}(i,j) $, $ \{\bar{y}(i,j)\}_{(i,j)} \in \mathcal{E} $. Specifically, we need to represent the likelihood of the active state of an edge $ (i, j) $ by associating the cost $ c = (1 - 2\bar{y}) $ of the edge $ (i, j) $ with the binary variable $ \tilde{y} $:
\begin{equation}
	\begin{split}
		\text{min}_{\tilde{y}} \quad & \tilde{y}(1 - 2\bar{y}) \\
		\sum_{(j,i) \in \mathcal{E} s.t. t_j < t_i} & \tilde{y}(j,i) \le 1 \quad \forall o_i \in \mathcal{V}\\
		\sum_{(i,k) \in \mathcal{E} s.t. t_i < t_k} & \tilde{y}(i,k) \le 1 \quad \forall o_i \in \mathcal{V} \\
	\end{split}	   
\end{equation}
where $ \tilde{y}(i,j) \in \{0, 1\}, \forall (i,j) \in \mathcal{E} $ is binary, it satisfies the constraint that, at most, one event edge is labeled 1 for each node's in-degree and out-degree. It is analogous to the conservation constraint used in the network flow problem \cite{networkflow_1993}. Ultimately, this objective is equivalent to minimizing the Euclidean distance $ \lVert \tilde{y} - \bar{y} \rVert_2 $. Solving the problem based on linear programming guarantees the algorithm obtains a globally optimal match in polynomial time \cite{pami_BerclazFTF11_KShort}.

\subsubsection{Optimization of The Learning Process Proof.} 
\begin{figure}[h]
	\centering
	\includegraphics[width=0.9\linewidth]{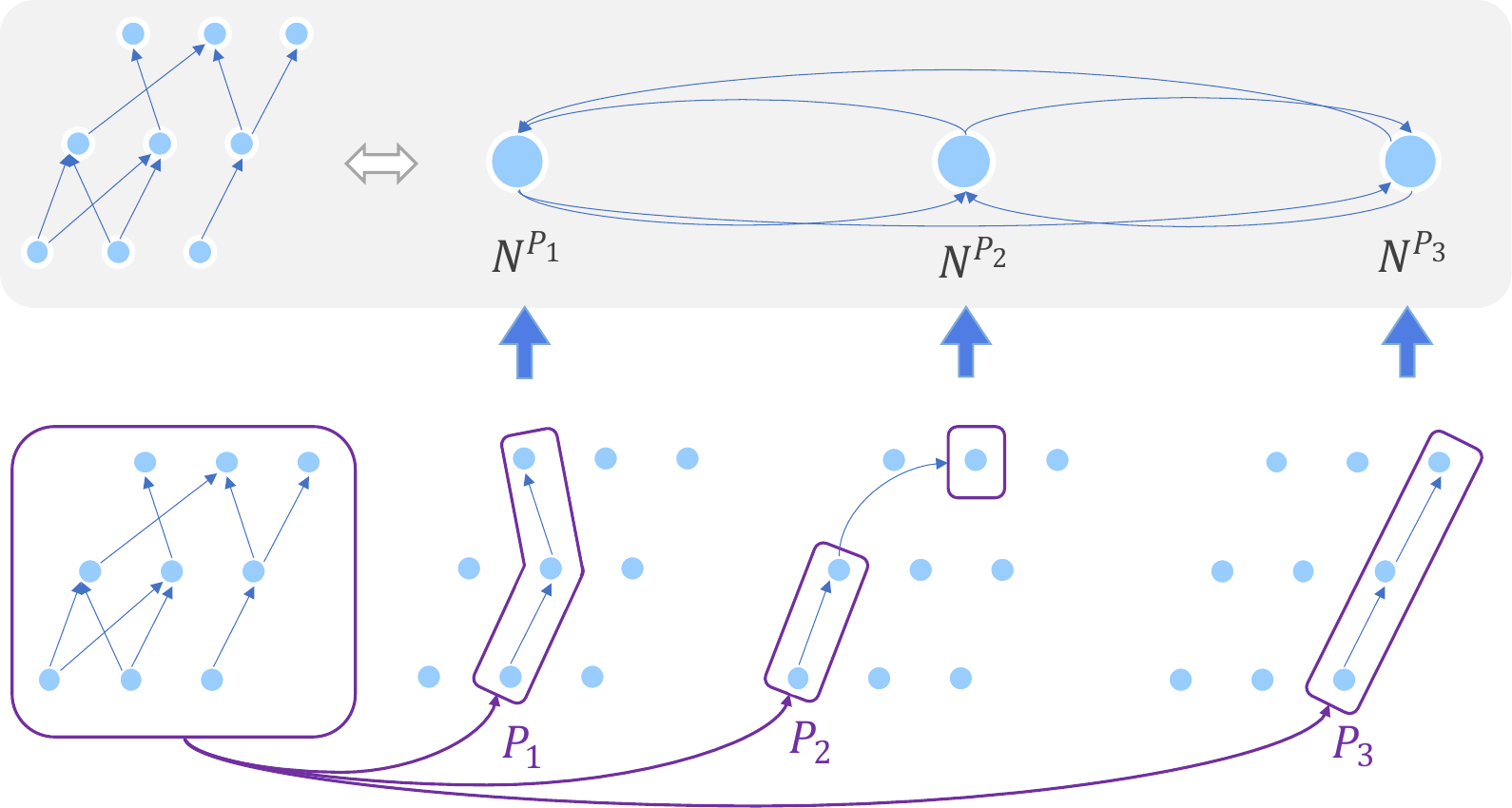}
	\caption{Branching strategy description. Search for the shortest plausible path from each vertex of the first frame to each vertex of the last frame of a video clip, and then represent it as a trajectory node, which is our ``branching'' representation.}
	\label{fig:branch}
\end{figure}

Figure \ref{fig:branch} depicts our optimization learning process. Indeed, our process can be explained using a branch-and-bound algorithm to approximate that increases the chances of finding tighter upper bounds earlier to achieve improvements. Our ultimate goal is to reduce the difficulty of the GNN to chunk the graph domain.

Although it is theoretically elegant, $ G^{local\_fully} $'s approach can only find the shortest optimal path of neighboring frames. It requires solving a total number $ |\mathcal{E}| \approx |\mathcal{V} \times k \times T/2| $ (each node chooses top-$ k $ nodes in the next frame to connect, a clip is divided into $ T/2 $ groups) shortest path problem in each iteration and is computationally more expensive than ``branch delimitation''.

The main idea of branch-and-bound is to iteratively subdivide the search space while using that result to tighten the upper and lower bounds. In this sense, instead of searching for the shortest possible path from each vertex of the first frame to each neighboring vertex, our $ G^{part} $ searches for the shortest plausible path from each vertex of the first frame of the video clip to each vertex of the last frame, which is then represented as a trajectory node (``branch'' representation). The GNN guarantees that our network flow is conserved, so the paths found can be considered the fittest. Thus, the delineation is done by the message-passing GNN. We then added all possible existing edges by creating fully connected edges to the trajectory nodes comprising the trusted paths. This ensures that all nodes with the same ID are reachable.

The short trajectory is a lower bound $ b(\cdot) $ on the globally optimal path if it is like an actual branch-and-bound algorithm. When both the first and last frames of an identity ID can be determined, an upper bound point location is found. However, it can only be recognized as a globally optimal path when the GNN confirms all other short paths within the segment. Therefore, the GNN must iteratively explore all frames before it has a comprehensive view of the trajectory.

The visual range of the GNN can be increased if one wants to find the upper bound point location of the trajectory that is tighter in time as early as possible. Instead of iteratively determining the $ G^{local\_fully} $ of the final trajectory, we input the node graph of the trajectory after complementary edges directly into the GNN to compute all the paths of a segment to confirm the trajectory's range earlier. Afterward, after learning from the GNN, we obtain the global optimal matching.

\section{B. Compared with Other Methods}
\label{sec:benchmark_results}
\begin{table}[!t]
	\centering
	\scalebox{0.6}{
		\begin{tabular}{lllll}
			\hline
			Method     & IDF1 $\uparrow$ & HOTA $\uparrow$ & MOTA $ \uparrow $ & IDS $\downarrow$ \\
			\hline
			\multicolumn{5}{c}{MOT17}\\
			\hline
			Tracktor \cite{iccv_BergmannML19_Tracktor}   & 55.1 & 44.8 & 56.3 & 1987 \\
			LPT \cite{cvpr_LiKR22_LPT}       & 57.7 & –    & 57.3 & 1424 \\
			MPNTrack \cite{cvpr_BrasoL20_MPNTrack}   & 61.7 & 49.0 & 58.8 & 1185 \\
			Lif\_T \cite{icml_HornakovaHRS20_Lif_T}     & 65.6 & 51.3 & 60.5 & 1189 \\
			ApLift \cite{iccv_HornakovaKSRRH21_ApLift}    & 65.6 & 51.1 & 60.5 & 1709 \\
			GMT \cite{cvpr_HeHWZ21_GMT}       & 65.9 & 51.2 & 60.2 & 1675 \\
			LPC MOT \cite{cvpr_DaiWCZHD21_LPC}   & 66.8 & 51.5 & 59.0 & 1122 \\
			SUSHI \cite{cvpr2023_Cetintas_SUSHI}         & 71.5 & 54.6 & 62.0 & 1041 \\
			\hline
			Ours       & \bf 72.0 & \bf 55.0 & \bf 62.4 & \bf 1033 \\ 
			\hline
			\multicolumn{5}{c}{MOT20}\\
			\hline
			Tracktor \cite{iccv_BergmannML19_Tracktor} & 52.7 & 42.1 & 52.6 & 1648 \\ 
			LPT \cite{cvpr_LiKR22_LPT}     & 53.5 & –    & 57.9 & 1827 \\ 
			ApLift \cite{iccv_HornakovaKSRRH21_ApLift}  & 56.5 & 46.6 & 58.9 & 2241 \\ 
			MPNTrack \cite{cvpr_BrasoL20_MPNTrack} & 59.1 & 46.8 & 57.6 & 1210 \\ 
			LPC MOT \cite{cvpr_DaiWCZHD21_LPC} & 62.5 & 49.0 & 56.3 & 1562 \\ 
			SUSHI \cite{cvpr2023_Cetintas_SUSHI} & 71.6 & 55.4 & 61.6 & \bf 1053 \\
			\hline
			Ours             & \bf 71.7 & \bf 55.6 & \bf 61.9 & 1060 \\ 
			\hline
	\end{tabular}}
	\caption{Test set results on MOT17 and MOT20 benchmark under public protocol.}
	\label{tab:mot17+20_pub}
\end{table}

\subsubsection{MOT17.} Under the public detection protocol, our model outperforms all published work using graph-based and hybrid graphs \cite{cvpr_LiKR22_LPT,iccv_HornakovaKSRRH21_ApLift,icml_HornakovaHRS20_Lif_T} (Table \ref{tab:mot17+20_pub}). All methods uniformly use Tracktor \cite{iccv_BergmannML19_Tracktor} preprocessing. Our method significantly outperforms them regarding long-term stable tracking, demonstrated by several metrics such as IDF1 and HOTA. Compared to MPNTrack, we improved 10.3 IDF1, 6.0 HOTA, and 3.6 MOTA. This suggests that the graph domains provided by our partially connected graphs and the learning approach facilitate more effective learning of GNNs. Compared to SUSHI, we improved 0.5 IDF1 and 0.4 HOTA, which suggests that time-continuous graph optimization has a positive significance for data association.

\subsubsection{MOT20.} In the dense scenario of MOT20, our detection setup is consistent with MOT17. We also achieved better performance compared to previous work. We significantly improved 12.6 IDF1, 8.8 HOTA, and 4.3 MOTA in the public setting. The reception field of the partial connectivity graph produces more interactions in the crowded scenario, with a slight advantage over the hierarchical full connectivity graph in several other metrics. These further highlights that our temporal information can provide valid cues.

\begin{table}[!t]
	\centering
	\scalebox{0.6}{
		\begin{tabular}{llccccc}
			\hline
			\# & Method     & FLOPs (G)  & FPS & VRAM Footprint (MiB) & $ E_{avg} $ & $ N_{avg} $ \\
			\hline 
			1 & $ G^{local\_fully} $  & 46.7 & 17.1 & 3689 & 256095 & 7482 \\
			2 & $ G^{part} $  & \bf 36.3 & \bf 18.9 & \bf 3523 & \bf 143603 & \bf 151 \\
			\hline
	\end{tabular}}
	\caption{Experiments on the MOT17 validation set.}
	\label{tab:cca_sup}
\end{table}

\subsubsection{Computational Complexity.} We compute the FLOPs and runtimes of SUSHI and the proposed method in Table \ref{tab:cca_sup}. Among them, FLOPs are the average FLOPs on val sequence (using thop library). VRAM footprint is the maximum VRAM occupied when reasoning over val sequences. Our $ G^{part} $ has a more reduced input size and fewer FLOPs overall. However, FPS is on par with SUSHI. The reason is that we ran the tracker mentioned in section 3.2.1 (see R3, A(4)-4) while building the graph. It is implemented in Python, slowing down the whole system. In addition, the gap for FPS exists because of GPU differences, NVIDIA A100 (SUSHI) and RTX 3090 (Ours), respectively. 

\section{C. Additional Details about CoNo-Link}
\label{sec:other_details}
\subsubsection{Current Graph Building Limitation.}
Given an input video with $ R $ frames, our first step is constructing a graph $ G $ for inter-frame object associations. It is mentioned above that edges of $ G $ introduced between neighboring frames represent the object's inter-relationships. We aim to guarantee the object associations between video frames that are far apart in time. However, constructing a fully connected graph $ G^{fully} $ within a clip could not be desirable. First, the edges of $ G^{fully} $ grow on the order of squares, and the longer the clip time, the greater the storage space requirement. Second, there is a labeling imbalance problem. The correct assumption is only a tiny fraction of the whole edges, thus affecting GNN learning.

Existing global multi-object tracking methods \cite{cvpr_BrasoL20_MPNTrack} usually mitigate this problem by rule-based edge pruning after building a fully connected graph for objects within a video clip. However, edge pruning will round off some actual edges. For this reason, the method \cite{cvpr2023_Cetintas_SUSHI} proposes a hierarchical completion of the full connectivity graph between neighboring frames. 
This solution is theoretically elegant, but (1) it restricts the receptive field of the GNN to neighboring frames, losing topological information about the trajectory. (2) it cannot correct some possible errors due to appearance similarity and occlusions.

\subsubsection{Association Cues.}
We obtain edge features from the trajectory graph and provide them to the multilayer perception MLP for classification. One of the main components is first the ReID-based appearance similarity. It is obtained by proposing an object patch using a pre-trained convolutional network. We compute the Euclidean distance between the trajectory's mean embeddings as the appearance distance $ \|f_{\mathrm{avg}}^{u}-f_{\mathrm{avg}}^{v}\|_{2} $.
For the identity preservation capability, feature representation can be more robust to confirm the reappearance of an object after occlusion.

In addition, we embed information such as temporal distance between nodes in the features to improve classification accuracy. It gives the basis for the nearest position of the trajectory in time.
Given the coordinates and time $ {(x_i, y_i, w_i, h_i, t_i)} $ of two trajectory nodes $ N^{traj_m} $ and $ N^{traj_n} $, define $ m $ as $ {(x_i, y_i, w_i, h_i, t_i)}^{m_{n_m}}_{i=m_1} $, $ n $ as $ {(x_i, y_i, w_i, h_i, t_i)}^{n_{n_n}}_{i=n_1} $. Assuming that $ N^{traj_m} $ precedes $ N^{traj_n} $, we compute the relative position and dimensional characteristics as:
\begin{equation}
	\Big(\frac{2(x_{n_{m}}-x_{n_{1}})}{h_{m_{n_{m}}}+h_{n_{1}}},\frac{2(y_{m_{n_{m}}}-y_{n_{1}})}{h_{m_{n_{m}}}+h_{n_{1}}},\mathrm{log}\,\frac{w_{m_{n_{m}}}}{w_{n_1}},\mathrm{log}\,\frac{h_{m_{n_{m}}}}{h_{n_1}}\Big) 
\end{equation}
We then connect this coordinate-based feature vector with the time difference $ t_{m_{n_m}} - t_{v_1} $ and the appearance appearance and feed it into the neural network GNN to obtain the initial edge embedding $ h^{(0)}_{(m,n)} $.

\subsubsection{Video Processing.} We are running on a video clip at 512 frames. We process the video as a sliding window and set the overlap between windows to 256 frames. In the overlapping windows section, we use two-part graph matching. For each pair of trajectories, $ T_{1} $ and $ T_{2} $, we can consider their IoUs as the association cost:
\begin{equation}
	c(T_{1},T_{2}): =\left\{\begin{array}{l l}{{1-IoU(T_{1},T_{2})}}&{{\mathrm{if~} \#(T_{1}\cap T_{2})>0}}\\ {{\infty}}&{{\mathrm{otherwise.}}}\end{array}\right.	
\end{equation}
where the IoU of two clips will be 1 when their trajectories are predicted to be the same and 0 when they do not overlap, and $ IoU(T_{1},T_{2}) =\frac{\#(T_{1}\cap T_{2})}{\#(T_{1}\cup T_{2})} $. In addition, non-overlapping tracks are strictly forbidden to match.

\subsubsection{Better Utilize the Properties of Networks.} We use NodeNet to define the edges of a partially connected graph $ G^{part} $. Our experiments found that the features containing global node information generated using Transformer can be well localized. Therefore, Top-k matching using NodeNet can effectively maintain GT edges while reducing the number of edges. Therefore, in our framework, an envelopment rate of 99.9 can be achieved by $ k = 5 $. In contrast, $ G^{fully} $ and $ G^{local\_fully} $ need $ k = 50 $ and $ k = 10 $ to fulfill the requirement with full connectivity and pruning, respectively. It indicates that the edges generated using neural network learning are more optimized than the hand-designed pruning method.

\section{D. Network Architecture}
\label{sec:architecture}
\begin{figure}[h]
	\centering
	\includegraphics[width=\linewidth]{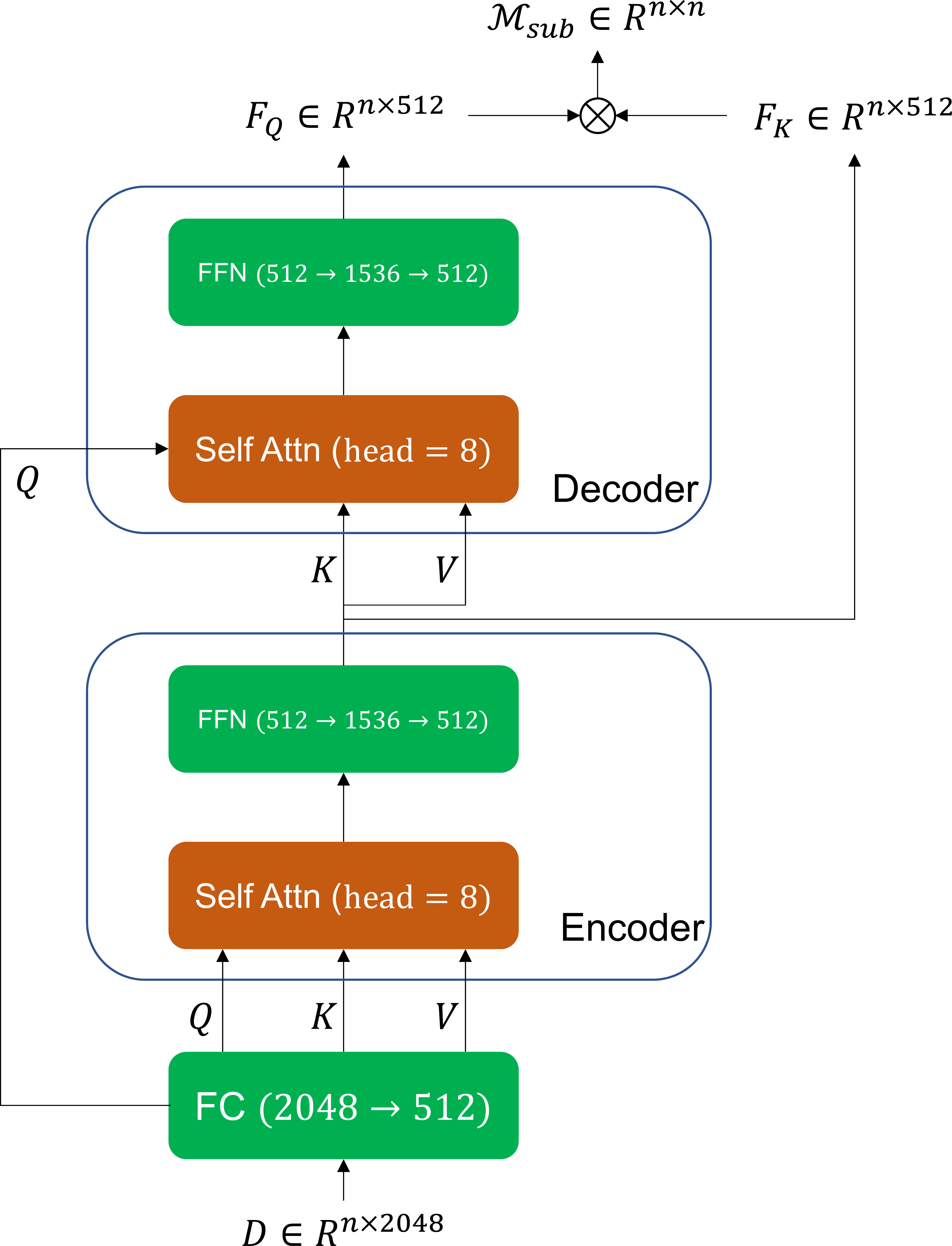}
	\caption{The detailed structure of NodeNet contains self-attn and cross-attn blocks. We list the data dimensions and multi-head numbers in parentheses.}
	\label{fig:nodenet}
\end{figure}
\begin{figure*}[h]
	\centering
	\includegraphics[width=0.7\linewidth]{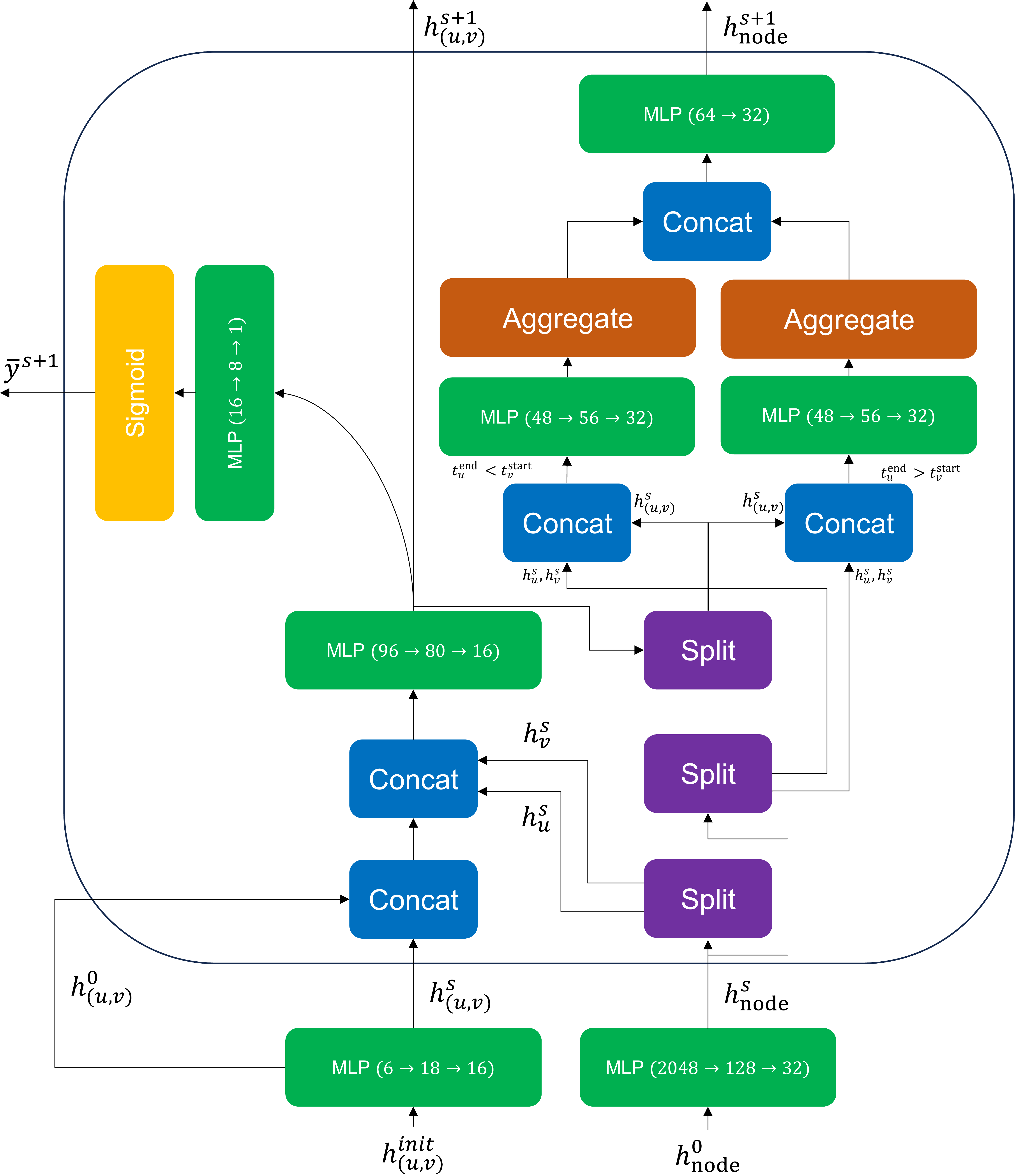}
	\caption{The detailed structure of the massage passing GNN contains various functional MLP components. We list the data dimensions in parentheses.}
	\label{fig:mpn}
\end{figure*}
\subsubsection{Update.} In this section, we detail a message passing GNN that considers temporal directionality. It updates edges by combining their node information. Then, it updates the nodes by aggregating the embeddings of the connected edges in past and future frames.
Given a graph, take as input our partially connected graph $ G^{part} $ and produce an embedding of higher-order node relations encoding the temporal information contained in it. We then use this for edge classification. Formally, for each graph $ \mathcal{G}^I = (\mathcal{V}^I, \mathcal{E}^I) $ in our iteration, we denote the feature embeddings of each node $ v \in \mathcal{V}^I $ and edge $ (u,v) \in \mathcal{E}^I $ as $ h^{(0)}_v \in R^{dv} $ and $ h^{(0)}_{(u,v)} \in R^{de} $, where $ dv $ and $ de $ are their respective dimensions. For each step $ s $, node $ v $, edge $ (u,v) $ is updated as follows.
\begin{equation}
	h_{(u,v)}^{(s)}=\mathrm{MLP}_{\mathrm{edge}}\left(\left[h_{u}^{(s-1)},\bar{h}_{(u,v)}^{(s-1)},h_{v}^{(s-1)}\right]\right). \qquad\quad	
\end{equation}

\begin{equation}
	m_{u \to v}^{(s)}=\left\{\begin{array}{l}{{\mathrm{MLP_{past}}\left([h_{u}^{(s-1)},{h}_{(u,v)}^{(s)},h_{v}^{(s-1)}]\right)\mathrm{if}\ t_{u}^{\mathrm{end}}<t_{v}^{\mathrm{start}}}}\\ {{\mathrm{MLP_{future}}\left([h_{u}^{(s-1)},{h}_{(u,v)}^{(s)},h_{v}^{(s-1)}]\right)\mathrm{else.}}}\end{array}\right.
\end{equation}

\begin{equation}
	h_\text{node}^{(s)}=\mathrm{MLP}_{\mathrm{node}} \left( \left[\sum_{u|t_{u}^{\mathrm{end}}<t_{v}^{\mathrm{start}}}m_{u\to v}^{(s)}, \sum_{u|t^{\mathrm{start}}<{t^{\mathrm{end}}}} m^{(s)}_{u\to v} \right] \right).
\end{equation}
where MLP denotes multilayer perceptron and $ [\cdot, \cdot] $ denotes concatenate, $ h_{(u,v)}^{(s)}:=[h_{(u,v)}^{(s)},h_{(u,v)}^{(0)}] $, $ t_{u}^{\mathrm{start}}$ and $t_{u}^{\mathrm{end}} $ denotes the timestamp of the track associated with node $ u \in \mathcal{V}^I $.

\subsubsection{Architecture.} The message-passing scheme described above relies on a set of lightweight multilayer perceptron. Figure \ref{fig:mpn} describes their exact architecture in detail, resulting in an embedding that enables high-precision edge classification. Specifically, it contains an MLP for initialized edge embeddings, an MLP for message passing, and an MLP for edge classification.
In addition, we have added a detailed structural diagram of NodeNet in Figure \ref{fig:nodenet}. Overall, our network's two most important modules are relatively lightweight and have less impact on the overall inference speed.

\section{E. Visualization}
\label{sec:visualization}
\begin{figure*}[h]
	\centering
	\includegraphics[width=0.9\linewidth]{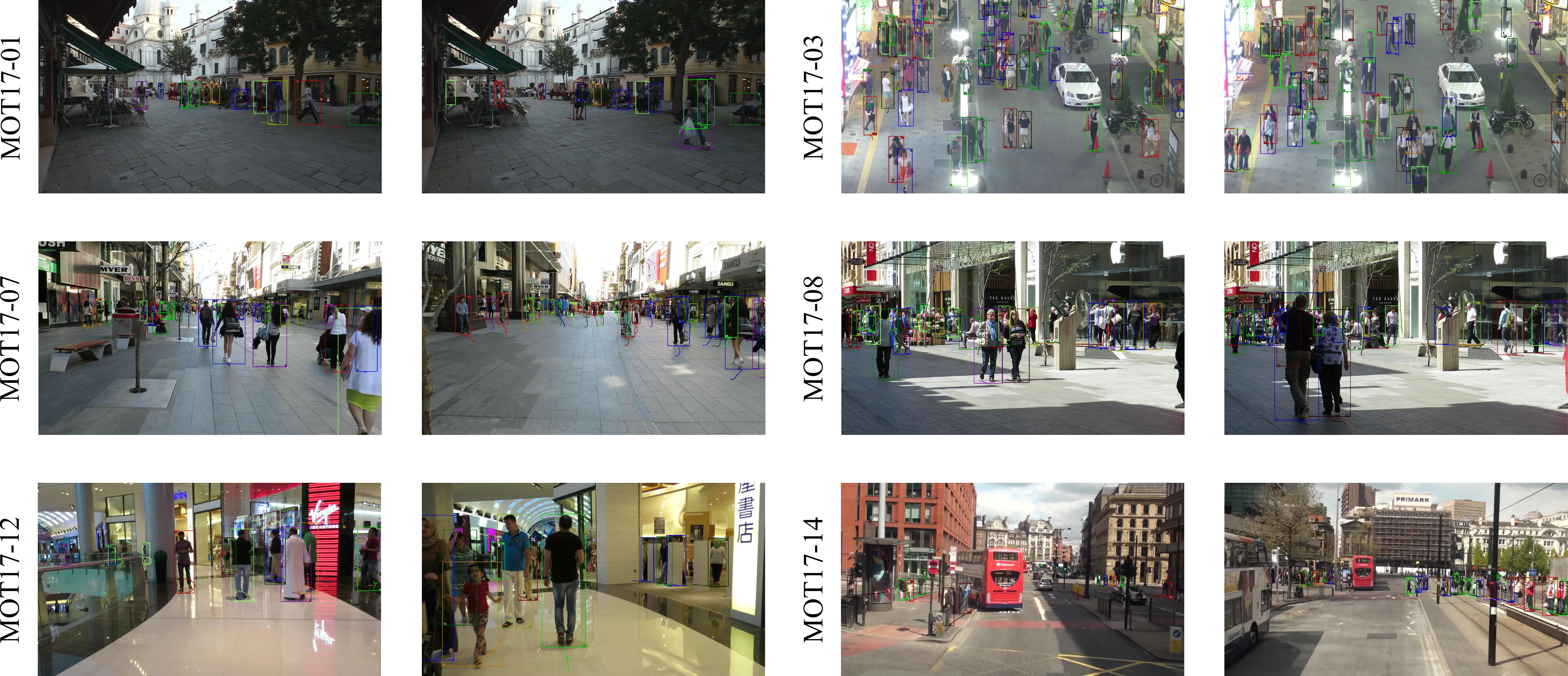}
	\caption{Qualitative result of CoNo-Link using private detections on MOT17\cite{MOT15_arxiv15,MOT16_arxiv16}. The bounding box color indicates the identity of each target. The curve under the bounding box is the trajectory of target.}
	\label{fig:mot17_vis}
\end{figure*}

\begin{figure*}[h]
	\centering
	\includegraphics[width=0.9\linewidth]{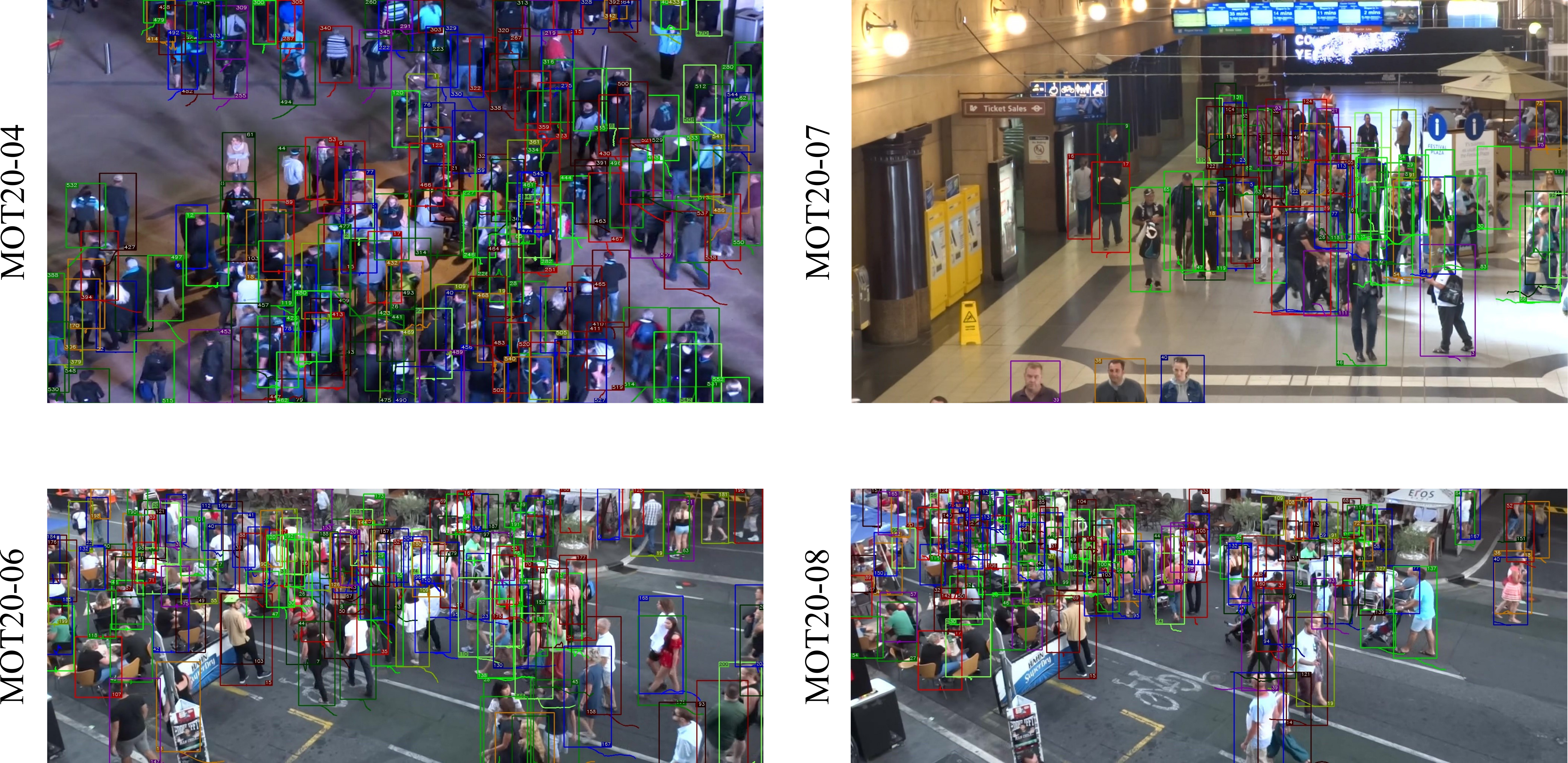}
	\caption{Qualitative result of CoNo-Link using private detections on MOT20\cite{Laura_MOT20}. The bounding box color indicates the identity of each target. The curve under the bounding box is the trajectory of target.}
	\label{fig:mot20_vis}
\end{figure*}

\begin{figure*}[h]
	\centering
	\includegraphics[width=0.9\linewidth]{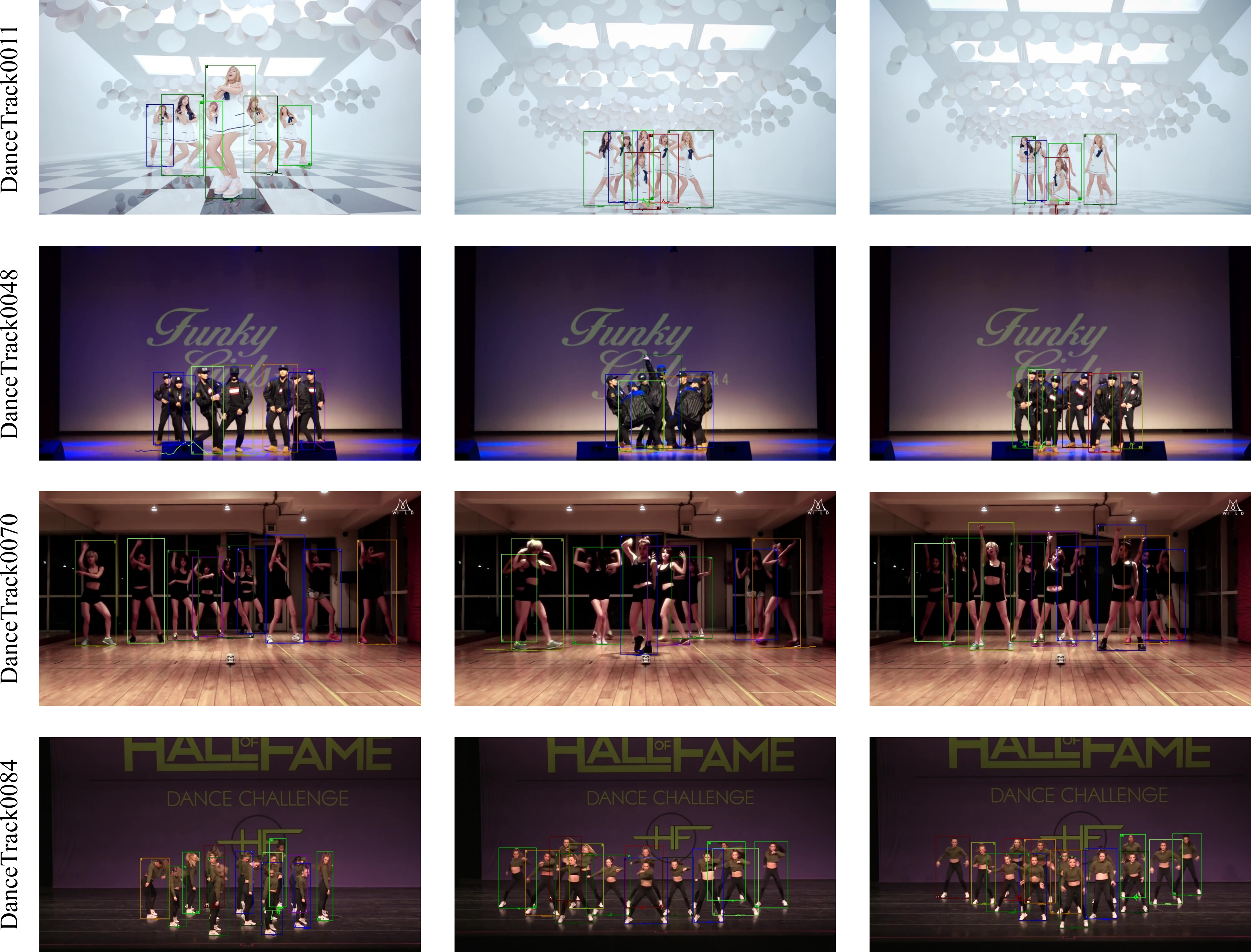}
	\caption{Qualitative result of CoNo-Link using private detections on DanceTrack\cite{dancetrack_cvpr22}. The bounding box color indicates the identity of each target. The curve under the bounding box is the trajectory of target.}
	\label{fig:dancetrack_vis}
\end{figure*}
In this section, we visualize the qualitative tracking results for each sequence in the MOT17, MOT20, and DanceTrack test sets, as shown in Figure \ref{fig:mot17_vis}, Figure \ref{fig:mot20_vis}, and Figure \ref{fig:dancetrack_vis}. Our tracker can perform robust tracking in multiple scenarios.

To analyze failure cases, we visualize our model on three datasets. We find that tracking the target after an long occlusion time leads to the non-existence of correctly associated objects within the Traj-Traj link. At this point, our method declines.

\end{document}